\documentclass[conference]{IEEEtran}
\IEEEoverridecommandlockouts
\usepackage{cite}
\usepackage{amsmath,amssymb,amsfonts}
\usepackage{algorithmic}
\usepackage{graphicx}
\usepackage{textcomp}
\usepackage{xcolor}
\usepackage{booktabs} 
\usepackage{multirow}
\usepackage{pifont}
\usepackage{tikz}
\usepackage{amsmath}
\usetikzlibrary{arrows.meta, positioning, shapes.geometric}
\usepackage{graphicx}
\usepackage{subcaption}
\usepackage[table]{xcolor}
\definecolor{gridblue}{HTML}{EAF2FA}
\usepackage{tabularx}
\usepackage{svg}

\newcommand{\cmark}{\ding{51}}
\newcommand{\xmark}{\ding{55}}
\newcommand{\std}[1]{{\scriptsize $\pm$ #1}}

\input{bib_ieee_abbr.def}

\def\BibTeX{{\rm B\kern-.05em{\sc i\kern-.025em b}\kern-.08em
    T\kern-.1667em\lower.7ex\hbox{E}\kern-.125emX}}
\begin{document}

\title{GridDriver: On-Board BEV-Based Language Planning and Reasoning\\

\thanks{$^{1}$Munich University of Applied Sciences, Intelligent Vehicles Lab (IVL), 80335 Munich, Germany
        {\tt\small intelligent-vehicles [at] hm.edu}}
}

\author{Katharina Winter$^{1}$, William Engel$^{1}$, Zain A. Alishammat$^{1}$, Fabian B. Flohr$^{1}$

}

\maketitle

\begin{abstract}
Bird’s-Eye-View (BEV)-conditioned large language models (LLMs) enable language-interactive driving, but onboard deployment requires efficient perception and continuous planning alongside autoregressive language generation. We present GridDriver, an efficient BEV-conditioned LLM framework combining trajectory prediction, structured driving indicators, natural-language navigation, driving explanations, and scene question answering. Lightweight spatio-temporal BEV perception and spatially structured tokens provide scene context to a shared LLM, while frame-wise scheduling interleaves trajectory prediction with asynchronous language generation. GridDriver achieves competitive performance on public benchmarks, reaching 87.9\% PDMS on NAVSIM and 57.6\% accuracy on nuScenes-QA with a shared planning and QA checkpoint. Spatial representation analyses demonstrate improved geometric information retention and trajectory prediction compared with Q-Former compression. The system achieves 32.7 ms standalone planning latency and 9.4 plans/s in an onboard ROS 2 pipeline. We further evaluate planning on our own data recordings and demonstrate open-loop onboard operation on a research vehicle. Across public benchmarks and real-world evaluations, we identify strengths and limitations of the BEV-based approach.
\end{abstract}

\begin{IEEEkeywords}
Autonomous Driving, Bird's-Eye-View Perception, Large Language Models, Language-Conditioned Planning, Real-Time Onboard Inference
\end{IEEEkeywords}

\section{Introduction}
\label{sec:intro}

Large language models (LLMs) have extended autonomous driving with natural-language instructions~\cite{winter2025bevdriver,zhou2026autovla,renz2025simlingo}, question answering~\cite{zhou2026opendrivevla,omnidrive,drivepi}, and driving descriptions~\cite{drivevlm}. These capabilities allow driving models to combine trajectory prediction with language understanding and generation. 
Grounding such models in the driving scene requires spatial information about road geometry, surrounding agents, and their motion from multiple camera views. 
Bird's-eye-view (BEV) representations fuse multi-camera observations into a shared ego-centric frame~\cite{bevformer}, providing spatial and temporal context aligned with trajectory prediction. This makes BEV a suitable interface for language-conditioned planning~\cite{rdadriver,drivepi,shao2024lmdrive,winter2025bevdriver}.

Moving BEV-conditioned LLM planning toward real-vehicle operation introduces additional requirements. Complex urban scenes demand rich spatial representations, while onboard computation requires efficient perception and inference. At the same time, the language capabilities motivating the use of an LLM should remain useful in practice, from natural-language navigation to passenger-facing driving explanations. We investigate these aspects through spatial representation analysis, joint planning and language evaluation, and exploratory onboard operation on a research vehicle.

\begin{figure}[t]
    \centering
    \resizebox{\columnwidth}{!}{%
        \includegraphics[width=0.9\columnwidth]{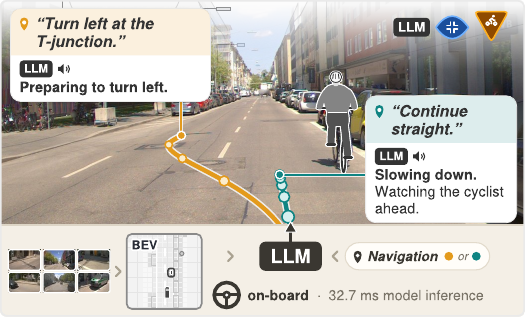}%
    }   
    \caption{GridDriver combines spatial BEV tokens and natural-language navigation in an LLM for trajectory planning and driving explanations, with an onboard implementation evaluated on a research vehicle.}
    \label{fig:eyecatcher}
\end{figure}

\begin{figure*}[htpb]
    \centering
    \includegraphics[
        width=\textwidth,
        keepaspectratio
    ]{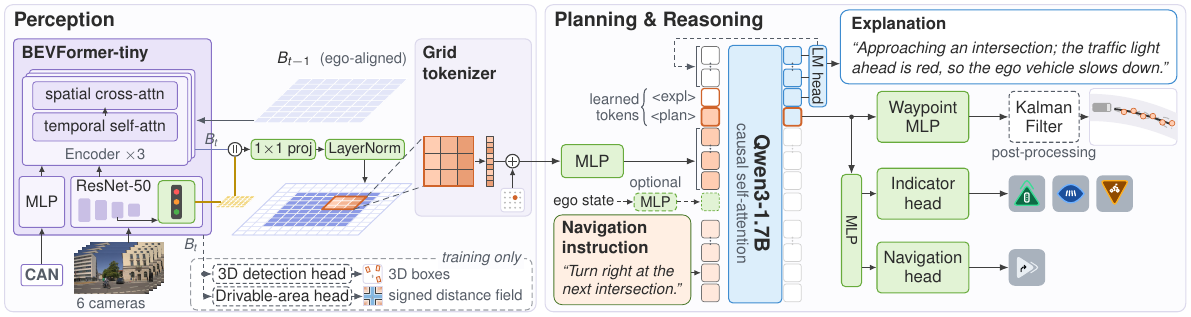}
    \caption{Framework architecture. BEVFormer-tiny~\cite{bevformer} encodes multi-camera images and CAN state into temporally aligned BEV features, augmented with traffic-light predictions. BEVFormer’s DETR-based 3D detection head and our drivable-area head support perception training. Grid tokens and navigation instructions are passed to Qwen3-1.7B, optionally with explicit ego-state. The LLM generates explanations, while the planning-token is decoded into future waypoints and indicators.}
    \label{fig:griddriver_architecture}
\end{figure*}

We propose GridDriver, a BEV-based LLM planner with natural-language input and output (see Figure~\ref{fig:eyecatcher}). We evaluate task-specific configurations on nuScenes, NAVSIM, and nuScenes-QA, and investigate deployment on a research vehicle. Our contributions are:
\begin{itemize}
    \item We demonstrate competitive language-conditioned trajectory prediction using a lightweight BEV-based LLM planner with temporally fused spatial tokens
    and direct waypoint regression.
    \item We extend the planner with free-form driving explanations and scene question answering, and investigate how language influences predicted trajectories.
    \item  We demonstrate real-time onboard BEV-based LLM inference with structured driving indicators and asynchronous language generation, and evaluate planning on geographically held-out urban recordings.
\end{itemize}
Code and checkpoints for the public datasets, and a documentation of the vehicle integration, will be made available.

\section{Related Work}
\label{sec:related_work}

\paragraph{Vision--Language Planning} 
Vision--language--action (VLA) driving models differ in how they represent visual observations and generate trajectories. Among front-view methods, SimLingo~\cite{renz2025simlingo} aligns driving and language predictions, SGDrive~\cite{li2026sgdrive} structures VLM features into a scene--agent--goal hierarchy for diffusion planning, and DriveVLA-W0~\cite{li2026drivevlaw0} uses world-model supervision. Multi-view methods include EMMA~\cite{emma}, which represents driving outputs as text, AutoVLA~\cite{zhou2026autovla}, which tokenizes feasible trajectories for autoregressive planning, and OpenDriveVLA~\cite{zhou2026opendrivevla}, which aligns 2D image and 3D instance-aware tokens with language. FSDrive~\cite{fsdrive} generates future visual frames as visual chain-of-thought, while DriveIK~\cite{park2026driveik} uses an LLM to predict future visual states, using current and predicted visual features for diffusion-based trajectory planning.
ReCogDrive~\cite{li2025recogdrive} combines VLM-guided diffusion planning with reinforcement learning.

\paragraph{BEV-Based LLM Planners} 
LMDrive~\cite{shao2024lmdrive} and BEVDriver~\cite{winter2025bevdriver} combine BEV features with language instructions using Q-Formers and GRU-based trajectory decoding, demonstrating closed-loop driving in CARLA. LMDrive encodes waypoint tokens, whereas BEVDriver directly leverages latent BEV features without precomputed waypoint tokens. Both use GRU-based trajectory decoding and demonstrate closed-loop driving in CARLA. DrivePI~\cite{drivepi} jointly predicts trajectories, occupancy, and language from BEV features, using diffusion-based action decoding. RDA-Driver~\cite{rdadriver} introduces patch-wise BEV tokens for LLM-based reasoning and autoregressive trajectory generation. Building on BEVDriver's planning paradigm and employing spatial tokenization similar to RDA-Driver, GridDriver combines temporally fused camera-based BEV features, direct waypoint regression, and asynchronous language generation for real-time onboard inference.

\paragraph{Vehicle Deployment and Language Interaction} 
Several language-enabled driving systems have been tested on vehicles. Alpamayo-R1~\cite{nvidia2025alpamayo} combines causal reasoning with diffusion-based trajectory planning, while DriveVLM-Dual~\cite{drivevlm} integrates VLM reasoning into a conventional driving stack. LINGO-2~\cite{wayve2024lingo2} jointly predicts driving trajectories and real-time language commentary on public roads. For human interaction, Cui et al.~\cite{cui2025onboard} translate spoken instructions into personalized control parameters, while Luo et al.~\cite{luo2025whatshappening} provide multimodal passenger-facing explanations. For efficient inference, AsyncDriver~\cite{chen2024asynchronous} asynchronously updates LLM guidance for a separate real-time planner, while OxyGen~\cite{li2026oxygen} combines KV-cache sharing with cross-frame batching for concurrent language and action generation. GridDriver instead interleaves BEV-conditioned planning with KV-cached explanation decoding in a shared LLM, enabling joint planning, QA, and language generation during public-road deployment.

\section{Method}
\label{sec:method}

GridDriver adopts the BEV-conditioned language planning framework established by BEVDriver~\cite{winter2025bevdriver}, integrating spatial BEV features with natural-language instructions for trajectory prediction. A modular perception stack employs BEVFormer~\cite{bevformer} to aggregate spatio-temporal features across camera views. As in RDADriver~\cite{rdadriver}, grid tokens preserve spatial scene structure while enabling efficient integration with a reasoning and planning decoder. The decoder jointly generates trajectory waypoints and textual scene descriptions. 

\paragraph{Perception} 
For efficient inference, we employ BEVFormer-tiny~\cite{bevformer} with a ResNet-50 backbone, three encoder layers, and half-resolution multi-view images. It produces a $50\times50\times256$ BEV representation covering $\pm51.2$\,m ($2.048$\,m per cell). Temporal information is incorporated through recurrent ego-motion-aligned BEV features. An ego-motion MLP injects an 18-dimensional vehicle-state vector into every BEV query to account for ego dynamics in the scene representation, coming from vehicle CAN bus or INS/IMU.
The encoder is trained for 3D detection with BEVFormer's DETR-style head and additionally fine-tuned with a drivable-area head, which regresses the signed distance rasterized from the HD map to the drivable-area boundary. Both heads shape the BEV only during encoder training.
A traffic-light head predicts four states (none, red, amber, green) for each of three directions using features from the encoder's frozen ResNet-50 backbone. The resulting 12 channels are concatenated with the BEV map and projected back to $256$ dimensions. After layer normalization and padding to $54\times54$, non-overlapping $3\times3$ patches are flattened, augmented with learned positional embeddings, and projected through a two-layer MLP ($2304\to2048\to2048$), yielding $18\times18=324$ BEV tokens.

\paragraph{Reasoning and Planning Decoder}
The LLM input is the navigation instruction, the 324 BEV tokens, a learned \texttt{<plan>} token and an optional explicit ego-state token. The latter is used in all NAVSIM configurations and ablated on nuScenes. In the NL variant, the four command slots are zeroed, so the route reaches the model only through the text. 
Under causal self-attention, every BEV token attends to the instruction and to the preceding cells, and the \texttt{<plan>} token, which attends to the whole sequence, is the read-out. 
A two-layer MLP maps the hidden state at the \texttt{<plan>} token to six waypoints $(x, y)$ in the ego frame at 0.5s intervals over 3s. For video visualizations and real-vehicle deployment, a causal Kalman filter smooths predicted trajectories to reduce frame-to-frame jitter. All quantitative planning evaluations use the raw, unfiltered predictions.
Icons and trajectory are read from the same hidden state. The icons therefore expose what the state that produces the plan encodes about the scene.
The LLM additionally generates free-form explanations of planning decisions. Within the same checkpoint, explanations can either follow the predicted plan, describing the decision, or precede it, allowing the plan to attend to the generated reasoning.

\section{Experimental Setup}
\label{sec:experiments}

\subsection{Datasets and Evaluation Protocols}
\textbf{nuScenes.} We evaluate open-loop planning on 6,019 validation keyframes (150 scenes) on L2 (m) and collision rate (\%). Following the GPT-Driver~\cite{gptdriver} ST-P3 scorer, where missing future steps contribute zero error and vehicle collisions are evaluated against future occupancy with ground-truth overlaps masked. We also evaluate nuScenes-QA on 83,337 questions from the same validation frames.

\textbf{NAVSIM.} We evaluate NAVSIM v1.1 navtest PDMS~\cite{navsim} and v2 navhard two-stage EPDMS~\cite{cao2025pseudo} using 4s trajectories. PDMS assesses collisions, drivable-area compliance, time-to-collision, progress, and comfort through non-reactive simulation with recorded agent trajectories. EPDMS adds traffic-rule, lane-keeping, and extended-comfort metrics. Second stage evaluates precomputed follow-up states, weighted by proximity to the first-stage endpoint, to approximate closed-loop behavior. Unlike L2 error, these scores assess driving quality without requiring exact imitation of the human trajectory.

\textbf{In-house urban dataset.} Our in-house dataset comprises approximately 251\,km of urban driving recorded in a single city recorded with six cameras, LiDAR, and a NovAtel INS at 10\,Hz, organized into 20s scenes with 2Hz keyframes. It contains 1,812 training scenes (4,240 manually annotated and 68,240 pseudo-labeled keyframes), 18 validation scenes (630 evaluated keyframes across three geographically held-out areas), and 268 additional unannotated scenes. We plan to release the dataset publicly in the future.

\paragraph{Navigation and Explanation Supervision}

\textbf{Natural-language navigation annotations} is generated for nuScenes, NAVSIM, and the in-house dataset from driven trajectories and map topology. Map-matched junction manoeuvres (left, right, straight, or U-turn) are combined with available landmarks and varied sentence templates, e.g. ``Take a left at the T-junction''. Unlike conventional direction commands, our annotations describe junction decisions rather than road curvature: bends without junctions receive ``Follow the road'', reducing direct cues about future trajectory shape. On nuScenes, 44\% of conventional left/right commands have no corresponding junction turn in our annotations, illustrating this distinction. We use these annotations for nuScenes and our in-house dataset.
\textbf{Explanation labels} are generated for nuScenes and in-house data from annotated 3D boxes, maps, and ego motion using rule-based templates, without human- or LLM-written text. Geometric rules identify interactions and causes of ego-motion. An independent reviewer checked 200 sampled validation frames per dataset against camera images. Labels were fully correct in 89.9\% of nuScenes frames and 90.4\% of in-house frames, with missed agents the main nuScenes error. On the same nuScenes frames, model-generated explanations without explicit ego-state input were fully correct in 80.0\%, with 87.0\% correct main statements. For NAVSIM, we use ReCogDrive's released QA-derived action and reason descriptions~\cite{li2025recogdrive}.

\subsection{Implementation and Training Details}

We use six cameras and Qwen3-1.7B with LoRA ($r=64$, $\alpha=128$) on attention and MLP projections. All BEVFormer-tiny encoders initialize from the official nuScenes-pretrained checkpoint, are fine-tuned per dataset, and frozen for planner training. nuScenes uses 6 encoder epochs with a drivable-area head; NAVSIM uses 9 epochs on navtrain detection labels, using six of eight cameras at half resolution. Drivable-area targets on nuScenes come from the map-expansion drivable and car-park layers, on NAVSIM from the scorer's own drivable-area polygons; supervision is only applied during training. In-house training uses up to 12 epochs without drivable-area supervision (missing map labels) on manually labeled training frames, selecting by validation mAP; ego-state comes from NovAtel INS. BEVFormer provides recurrent temporal fusion, restricted on NAVSIM to four 2Hz frames (current + 3 past). With cached BEV features, we train the grid tokenizer, ego-state MLP, waypoint head, and LoRA adapters for 30 epochs using mixed-precision AdamW (peak LR $10^{-4}$, 2,000 warm-up steps, cosine decay, weight decay 0.06, batch size 64). Training runs on NVIDIA A40 and A100 GPUs. Encoder fine-tuning takes about 28 A40 GPU-hours on nuScenes. With cached BEV features, one 30-epoch planner run takes about 12 A100 GPU-hours on nuScenes (2 GPUs, ~6 h) and about 43 A100 GPU-hours on NAVSIM (4 GPUs, ~11 h).
We predict ego-frame waypoints at 0.5\,s intervals: six (3\,s) for nuScenes and the in-house dataset and eight (4\,s) for NAVSIM. NAVSIM uses ego status and waypoint headings are obtained by central differences of predicted positions. The official scorer tracks these poses using LQR and a kinematic bicycle model, without test-time simulator access. We report navtest PDMS (v1.1) and non-reactive EPDMS (v2).

Evaluations average three seeds unless stated otherwise.
nuScenes uses seeds 42,1,2 with checkpoints selected by minimum validation planning L2 (23,23,18, also for QA). NAVSIM trains on all navtrain with seeds 42,43,44, using the final epoch (29). In-house evaluation uses seeds 42,2,1 with best val epochs 21,22,25 with ego-state and 26,8,15 without planner ego-state respectively. Benchmark evaluations and onboard inference use post-plan, language interventions use pre-plan explanations.

The training objective combines waypoint regression with optional language, icon and navigation supervision:
\begin{equation}
\mathcal{L}=\mathcal{L}_{\mathrm{wp}}+\lambda_{\mathrm{text}}\mathcal{L}_{\mathrm{text}}
+\lambda_{\mathrm{icon}}\mathcal{L}_{\mathrm{icon}}+\lambda_{\mathrm{nav}}\mathcal{L}_{\mathrm{nav}},\label{eq:loss}
\end{equation} 
where $\mathcal{L}_{\mathrm{wp}}$ is the mean L1 waypoint error, $\mathcal{L}_{\mathrm{text}}$ is next-token cross-entropy for teacher-forced explanations, and $\mathcal{L}_{\mathrm{icon}}$ is binary cross-entropy over 18 driving indicator icons covering road conditions, following, vehicle states, watching, and yielding. $\mathcal{L}_{\mathrm{nav}}$ is class-weighted cross-entropy over maneuver (none, straight, left, right, U-turn) and phase (none, approaching, turning). Their combinations determine four navigation display icons: preparing-left/right and turning-left/right. The indicator and navigation heads are linear classifiers on a shared two-layer trunk that reads the same LLM hidden state as the waypoint head, adding no decoding step. We set $\lambda_{\mathrm{text}}=\lambda_{\mathrm{icon}}=1$ and $\lambda_{\mathrm{nav}}=0.2$ when labels are available and 0 otherwise. We sample pre-plan and post-plan explanations equally. indicator and navigation losses apply only to post-plan samples, preventing target leakage from the explanation. Explanations average 9 tokens (maximum 32).

\subsection{Baselines and Comparability}

Baselines primarily comprise published VLA models, alongside non-VLA methods such as DiffusionDrive~\cite{liao2025diffusiondrive} and SimWAM~\cite{zhao2026simwam}, with fixed-policy configurations with varying training data, backbones (0.5B--8B for language models), and evaluation protocols. NAVSIM v2 component scores for DiffusionDrive, ReCogDrive, SGDrive, and DriveVLA-W0 follow SimWAM~\cite{zhao2026simwam}. $^*$ marks ST-P3 variants of unverified equivalence. GridDriver uses a nuScenes-pretrained BEV encoder and benchmark-specific imitation learning with no separate large-scale driving-VQA pretraining of the LLM or test-time adaptation. 
The NL evaluations replace standard direction commands with junction-aware, map-derived instructions, while retaining the same planning architecture.

\subsection{Experimental Vehicle Setup}
We integrate GridDriver into a ROS 2 Humble pipeline on a research vehicle equipped with six surround cameras and a NovAtel GNSS/INS. Camera images, odometry, and IMU measurements are approximately synchronized (80ms tolerance) and combined with navigation instructions from Mapbox. The system publishes timestamped 3D object detections, traffic-light probabilities, six future waypoints as a ROS path, and 18 driving indicator predictions alongside maneuver and phase predictions for the navigation icons. Natural-language explanations are generated asynchronously, with completed sentences displayed for at least 2s to reduce flicker. Icon events independently trigger pre-rendered speech. Sensor timestamps, ego-state features, and navigation inputs are logged for reproducibility. For onboard visualization, predicted trajectories are smoothed using a causal Kalman filter, which does not modify the model's predictions.
The BEV encoder and LLM planner execute separately on two RTX 4090 GPUs, with preprocessing pipelined alongside inference. A single-frame queue and stale-frame dropping prevent latency accumulation. The scheduling of trajectory prediction and language generation is detailed in Section~\ref{sec:inference}.

\section{Results}
\label{sec:results}

\begin{table}[t]
\vspace{1.2mm}
\centering
\caption{Open-loop nuScenes planning (mean 1--3s L2 and collision rate). Planner Ego denotes explicit ego-state input to the planner; GridDriver retains BEVFormer CAN conditioning. External scores may differ in evaluation protocol; $^*$ marks unverified ST-P3 variants. GridDriver: cmd: standard navigation command, NL: natural-language instruction.}
\label{tab:nuscenes}
\small
\renewcommand{\arraystretch}{0.90}
\setlength{\tabcolsep}{3.5pt}
\begin{tabular}{@{}lcll@{}}
\toprule
Method & Planner Ego & avg. L2 $\downarrow$ & avg. Col. $\downarrow$ \\
\midrule
FSDrive$^{*}$~\cite{fsdrive} & & 0.53 & 0.17 \\
DrivePI$^{*}$~\cite{drivepi} & & 0.49 & 0.38 \\
FSDrive$^{*}$~\cite{fsdrive} & \cmark & 0.28 & 0.10 \\
AutoVLA (CoT)$^{*}$~\cite{zhou2026autovla} & \cmark & 0.40 & 0.20 \\
RDADriver~\cite{rdadriver} & \cmark & 0.40 & 0.10 \\
DrivePI$^*$~\cite{drivepi} & \cmark & 0.40 & 0.11 \\
DriveIK-0.5B$^*$~\cite{park2026driveik} & \cmark & 0.06 & 0.09 \\
\midrule
Constant Velocity & \cmark & 0.61 & 0.22 \\
Ego-MLP & \cmark & 0.32 & 0.11 \\
OpenDriveVLA-0.5B~\cite{zhou2026opendrivevla} & & 0.75 & 0.24 \\
OpenDriveVLA-0.5B~\cite{zhou2026opendrivevla} & \cmark & 0.35 & 0.09 \\
OpenDriveVLA-3B~\cite{zhou2026opendrivevla} & \cmark & 0.33 & 0.10 \\
\rowcolor{gridblue}
\textbf{GridDriver} (ours) (cmd) & & 0.47 $\pm$ 0.01 & 0.15 $\pm$ 0.01  \\
\rowcolor{gridblue}
\textbf{GridDriver} (ours) (NL) & & 0.47 $\pm$ 0.00 & 0.15 $\pm$ 0.01  \\
\rowcolor{gridblue}
\textbf{GridDriver}(ours) (cmd) & \cmark & 0.33 $\pm$ 0.01 & 0.10 $\pm$ 0.00 \\
\rowcolor{gridblue}
\textbf{GridDriver}(ours) (NL) & \cmark & 0.33 $\pm$ 0.01 & 0.11 $\pm$ 0.01 \\
\bottomrule
\end{tabular}
\end{table}


\begin{table}[t]
\centering
\caption{NAVSIM v1.1 navtest PDMS (\% $\uparrow$). NC: no at-fault collisions; DAC: drivable area compliance; TTC: time to collision; EP: ego progress; C: comfort. GridDriver: cmd: standard navigation command, NL: natural-language instruction. PDMS cmd std.: 0.15\,pp. PDMS NL std.: 0.06\,pp}
\label{tab:navsim}
\small
\renewcommand{\arraystretch}{0.95}
\setlength{\tabcolsep}{3pt}
\begin{tabularx}{\columnwidth}{@{}Xrrrrrr@{}}
\toprule
Method & NC & DAC & TTC & EP & C & PDMS \\
\midrule
Constant Velocity & 68.0 & 57.8 & 50.0 & 19.4 & 100.0 & 20.7 \\
DiffusionDrive~\cite{liao2025diffusiondrive} & 98.2 & 96.2 & 94.7 & 82.2 & 100.0 & 88.1 \\
FSDrive~\cite{fsdrive} & 98.2 & 93.8 & 93.3 & 80.1 & 99.9 & 85.1 \\
AutoVLA~\cite{zhou2026autovla} & 96.9 & 92.4 & 88.1 & 75.8 & 99.9 & 80.5 \\
ReCogDrive-IL~\cite{li2025recogdrive} & 98.1 & 94.7 & 94.2 & 80.9 & 100.0 & 86.5 \\
SGDrive (SFT)~\cite{li2026sgdrive} & 98.6 & 95.1 & 95.4 & 81.2 & 100.0 & 87.4 \\
DriveVLA-W0~\cite{li2026drivevlaw0} & 98.7 & 96.2 & 95.5 & 82.2 & 100.0 & 88.4 \\
DriveIK-SFT~\cite{park2026driveik} & 99.1 & 96.2 & 99.0 & 83.4 & 100.0 & 90.3 \\
\rowcolor{gridblue}
\textbf{GridDriver}(ours) (cmd)~\ & 98.5 & 95.7 & 94.9 & 82.0 & 100.0 & 87.9 \\
\rowcolor{gridblue}
\textbf{GridDriver}(ours) (NL)~\ & 98.3 & 95.0 & 94.3 & 82.0 & 99.9 & 87.1 \\

\bottomrule
\end{tabularx}
\end{table}

\begin{table*}[t]
\vspace{1.2mm}
\centering
\caption{NAVSIM v2 navhard two-stage EPDMS (\% $\uparrow$). Each component is Stage 1/2; NC: no at-fault collisions, DAC: drivable area compliance, DDC: driving direction compliance, TLC: traffic light compliance, EP: ego progress, TTC: time to collision, LK: lane keeping, HC: history comfort, EC: extended comfort. EPDMS std.: 1.1\,pp.}
\label{tab:navsim_epdms}
\small
\renewcommand{\arraystretch}{1.08}
\setlength{\tabcolsep}{3pt}
\begin{tabular}{@{}l*{9}{c}r@{}}
\toprule
Method & NC & DAC & DDC & TLC & EP & TTC & LK & HC & EC & EPDMS \\
\midrule
Constant velocity & 88.9/83.2 & 42.9/59.1 & 70.7/76.5 & 99.3/98.1 & 77.5/71.4 & 87.3/81.1 & 78.7/48.0 & 97.1/97.2 & 60.4/62.0 & 11.5 \\
DiffusionDrive~\cite{liao2025diffusiondrive} & 96.8/80.1 & 86.0/72.8 & 98.8/84.4 & 99.3/98.4 & 84.0/85.9 & 95.8/76.6 & 96.7/46.4 & 97.6/96.3 & 79.6/72.8 & 27.5 \\
SimWAM~\cite{zhao2026simwam} & 98.0/81.8 & 92.0/78.6 & 99.7/87.3 & 99.6/98.4 & 83.8/86.3 & 96.2/78.6 & 97.3/49.5 & 97.8/96.3 & 71.6/69.9 & 37.6 \\
DriveVLA-W0~\cite{li2026drivevlaw0}
 & 96.8/76.8 & 83.3/64.3 & 99.0/79.9 & 99.6/98.3 & 84.6/89.2 & 95.3/75.0 & 96.4/46.8 & 97.6/95.8 & 78.2/53.1 & 24.4 \\
SGDrive-IL~\cite{li2026sgdrive}
& 95.8/79.4 & 87.6/65.4 & 97.8/79.1 & 99.8/98.9 & 84.4/88.9 & 94.7/75.3 & 92.9/42.7 & 97.8/96.4 & 28.9/29.6 & 25.5 \\
ReCogDrive~\cite{li2025recogdrive}-IL & 96.4/80.2 & 78.9/65.0 & 98.7/82.4 & 99.8/98.7 & 82.6/85.2 & 95.6/76.9 & 94.4/43.8 & 97.6/96.6 & 74.2/71.8 & 25.7 \\
\rowcolor{gridblue}
\textbf{GridDriver} (ours) (cmd)~\ & 96.6/80.5 & 84.2/71.0 & 99.1/84.3 & 99.8/98.4 & 84.0/85.2 & 95.1/77.7 & 94.9/47.0 & 97.8/96.2 & 79.6/71.9 & 28.2 \\
\bottomrule
\end{tabular}
\end{table*}

\subsection{Public Benchmark Results}

\paragraph{Planning}
Without an additional explicit ego-state token in the LLM, GridDriver achieves 0.47\,m mean L2 error and 0.15\% collision rate on nuScenes (Table~\ref{tab:nuscenes}), compared with 0.75\,m and 0.24\% reported for OpenDriveVLA-0.5B when its explicit ego input is ablated. Both use BEVFormer-based perception, which supports ego-motion alignment and CAN-bus conditioning. Thus, removing explicit planner ego-state does not guarantee ego-state-free BEV features, which may introduce ego-state shortcuts~\cite{li2024ego}. Adding an explicit ego-state token improves GridDriver to 0.33\,m and 0.10\%, comparable to OpenDriveVLA-3B. 

On NAVSIM v1, GridDriver achieves $87.9\%\pm0.15\%$ PDMS (Table~\ref{tab:navsim}) and on v2, $28.2\pm1.1\%$ EPDMS (Table~\ref{tab:navsim_epdms}), remaining competitive with contemporary VLA planners without RL fine-tuning.
Junction-aware natural-language instructions achieve comparable planning performance to standard commands on nuScenes (0.33 m L2 for both) and NAVSIM (87.1\% vs. 87.9\% PDMS).
These results indicate that junction-aware natural-language instructions can replace standard direction commands with little impact on planning performance, but do not establish generalization to arbitrary passenger instructions.

\paragraph{nuScenes-QA}
QA prompts append a dataset question to the navigation instruction, e.g., ``Make a left turn at the traffic light. Question: What number of cars are to the front left of the moving car?'' 
GridDriver achieves 57.6\% QA accuracy (Table~\ref{tab:nuscenes_qa}), below OpenDriveVLA-0.5B (58.4\%) and exceeding DriveIK (56.4\%). Unlike OpenDriveVLA, which reports QA before its final planning-training stage~\cite{zhou2026opendrivevla}, GridDriver retains 0.33\,m planning L2 and 0.11\% collision rate with the same checkpoint. Limited BEV coverage ($\pm51.2$\,m) complicates counting, as approximately 70\% of all questions and 74\% of counting questions involve objects outside this region.
Overall, the retained nuScenes-QA capability motivates extending onboard interaction beyond driving explanations to passenger-initiated scene questions.

\begin{table}[t]
\centering
\caption{nuScenes-QA val. accuracy (\%). Ext.: existence, Obj.: object, Sts.: status, Cnt.: counting, Cmp.: comparison, P: same checkpoint for QA and planning. Accuracy std.: 0.6.}
\label{tab:nuscenes_qa}
\small
\renewcommand{\arraystretch}{1.0}
\setlength{\tabcolsep}{1.2pt}
\begin{tabularx}{\columnwidth}{
@{}>{\raggedright\arraybackslash}Xcrrrrrr@{}}
\toprule
Method & P & Ext. & Obj. & Sts. & Cnt. & Cmp. & Acc. \\
\midrule
OpenDriveVLA-0.5B~\cite{zhou2026opendrivevla}
& \xmark & 83.9 & 50.2 & 57.0 & 22.0 & 68.4 & 58.4 \\
OpenDriveVLA-3B~\cite{zhou2026opendrivevla}
& \xmark & 84.0 & 50.3 & 56.9 & 22.3 & 68.5 & 58.5 \\
DriveIK-0.5B~\cite{park2026driveik} & \cmark & 83.7 & 47.1 & 52.8 & 19.5 & 67.3 & 56.4 \\
DrivePI~\cite{drivepi}
& \cmark & 85.3 & 57.5 & 59.1 & 22.4 & 68.3 & 60.7 \\
\rowcolor{gridblue}
\textbf{GridDriver} (ours) (NL)
& \cmark & 83.3 & 51.4 & 54.6 & 20.6 & 66.9 & 57.6 \\
\bottomrule
\end{tabularx}
\end{table}

\subsection{In-House Open Loop Results}
\label{sec:munich-results}
\paragraph{Qualitative Results}
\label{par:qualitative}

\begin{figure*}[t]
    \centering

    \begin{subfigure}[b]{0.33\textwidth}
        \centering
        \includegraphics[width=\linewidth]{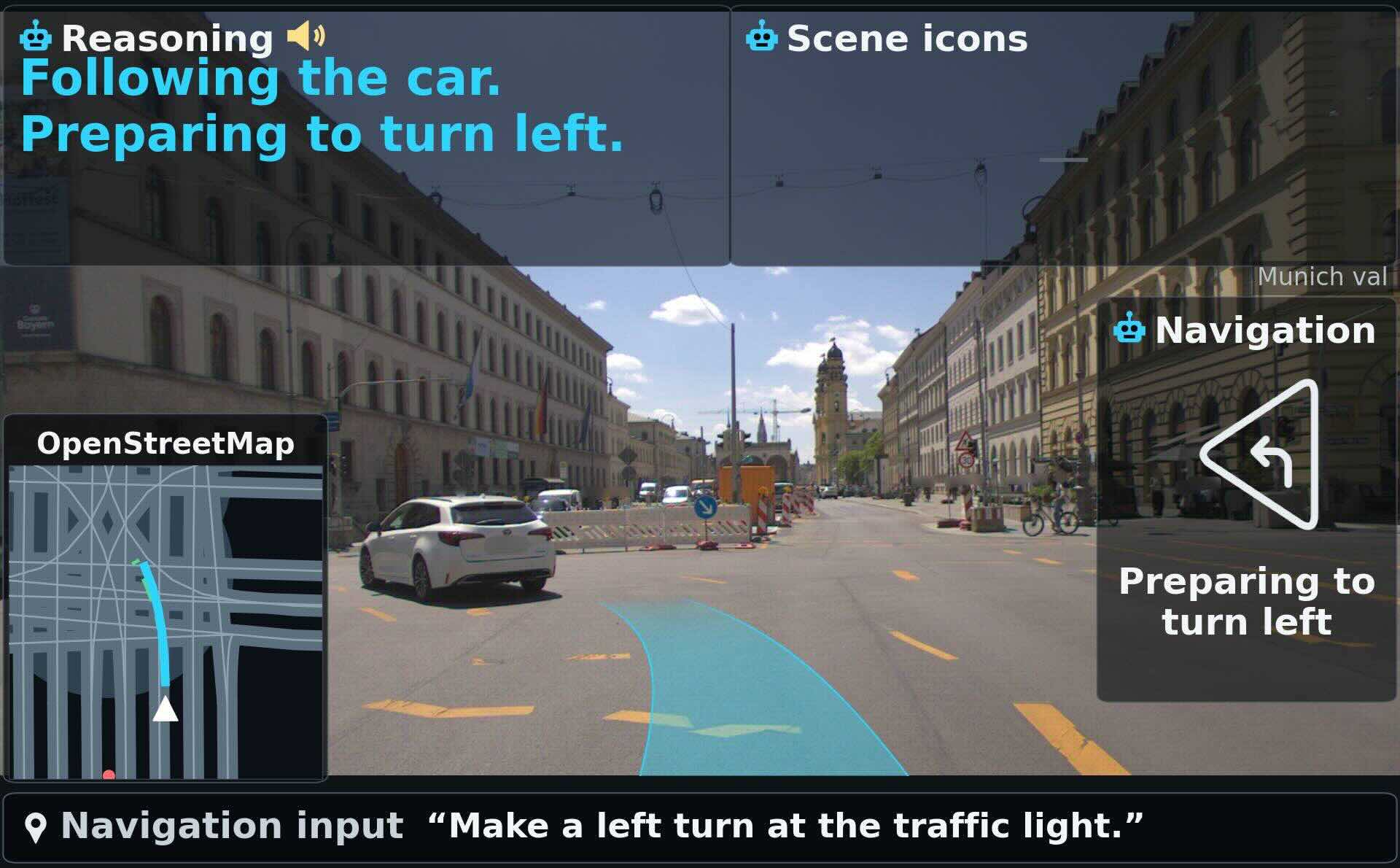}
        \caption{Prepare Turn.}
        \label{fig:a}
    \end{subfigure}%
    \hfill
    \begin{subfigure}[b]{0.33\textwidth}
        \centering
        \includegraphics[width=\linewidth]{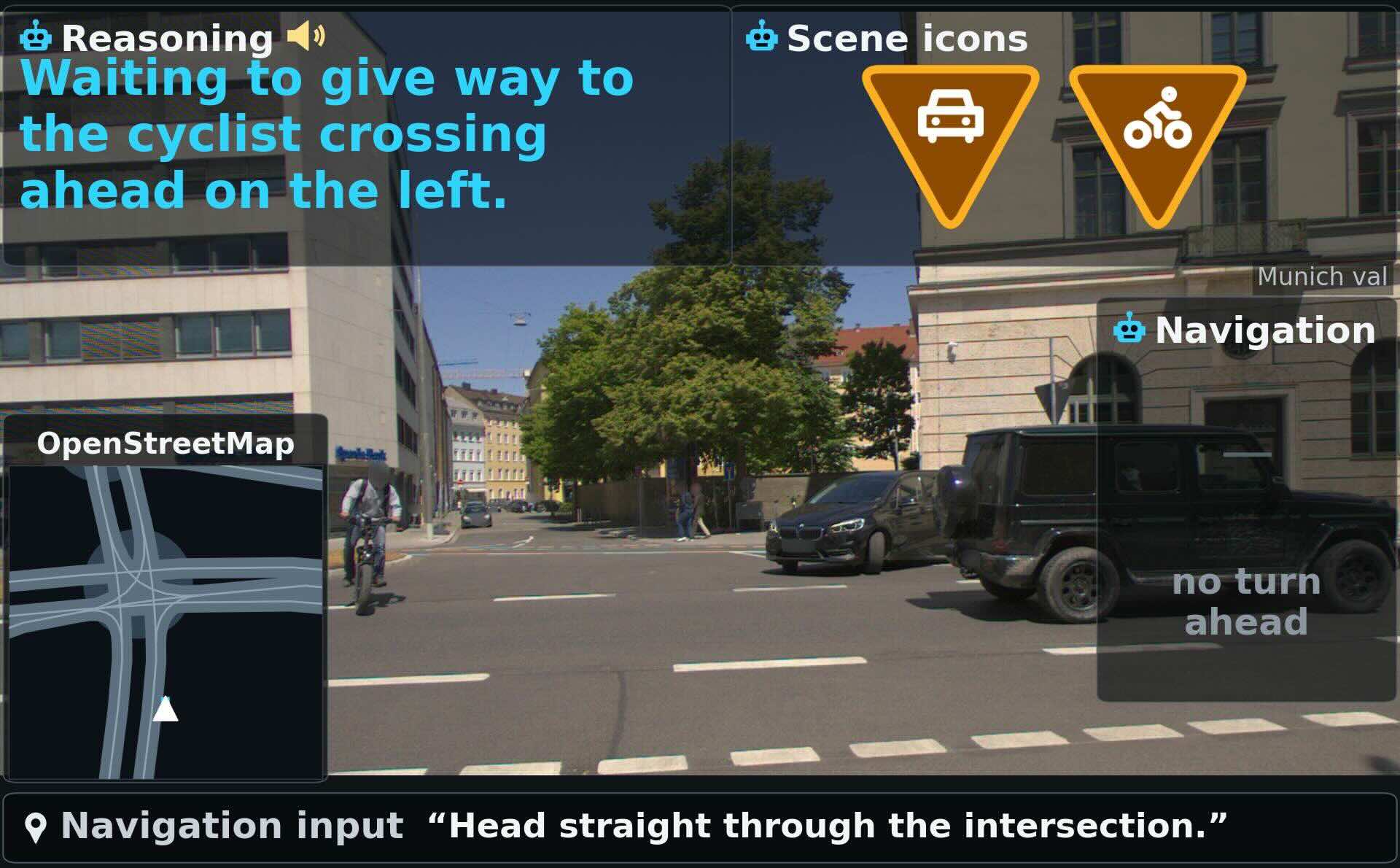}
    \caption{Yielding at intersection.}
        \label{fig:b}
    \end{subfigure}%
    \hfill
    \begin{subfigure}[b]{0.33\textwidth}
        \centering
        \includegraphics[width=\linewidth]{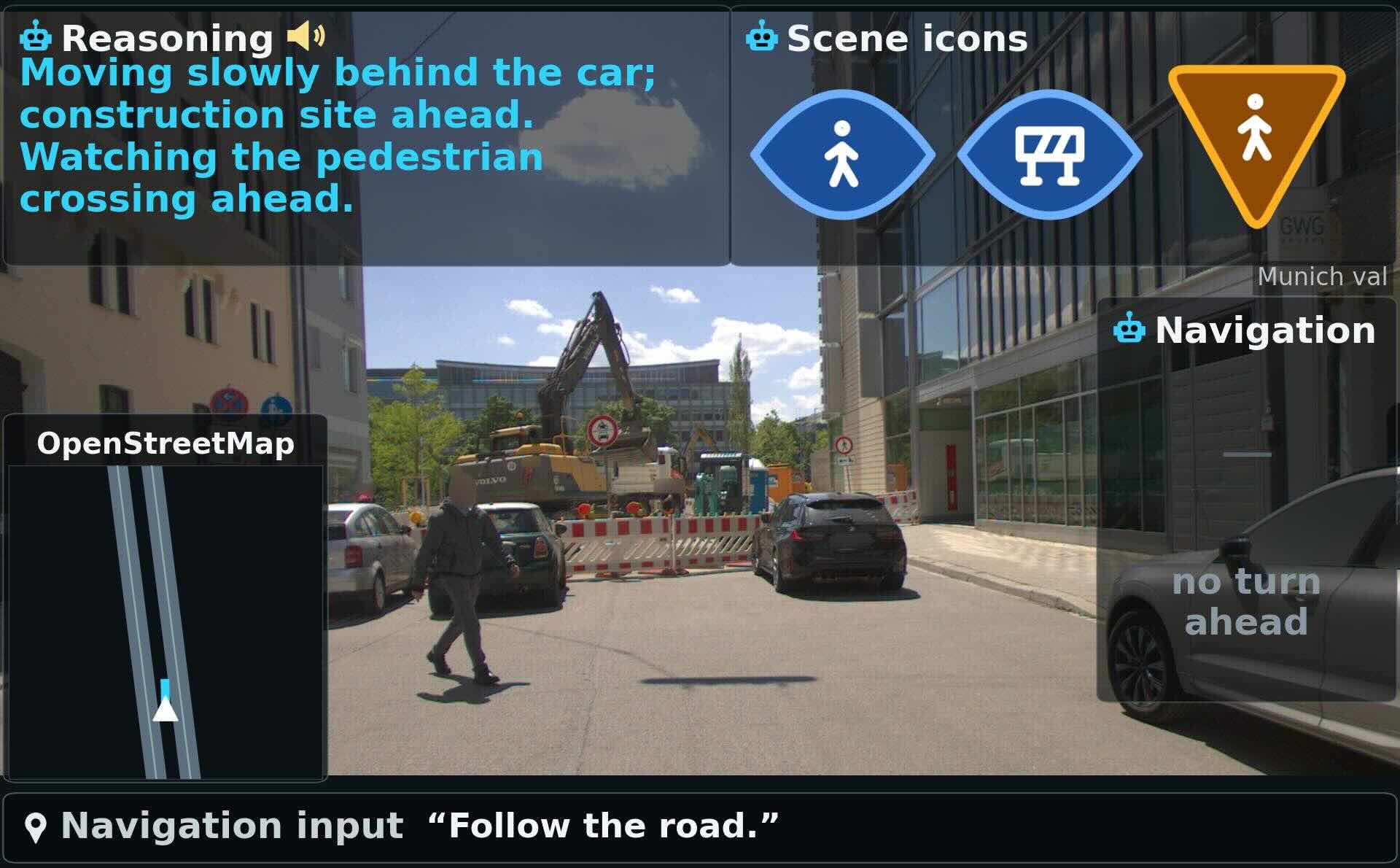}
        \caption{Failure Case: Wrong cause.}
        \label{fig:c}
    \end{subfigure}%
    \hfill

    \caption{Qualitative examples. Interface showing model predictions across three driving scenarios (anonymized).}
    \label{fig:qualitative-results}
\end{figure*}

Figure~\ref{fig:qualitative-results} presents qualitative examples from in-house validation videos of the with-ego checkpoint (seed 42) on navigation instructions (bottom left), predicted trajectories (blue band) overlaid on an OpenStreetMap top-down view, model reasoning (top left), predicted driving indicators (top right), and navigation icons (right). The examples illustrate anticipating a left turn (\ref{fig:a}) and yielding to a cyclist and a car at an intersection (\ref{fig:b}). In the failure case (\ref{fig:c}), the model incorrectly attributes slowing down near a construction-related dead end to a moving vehicle, although the surrounding cars are parked. Although the planned behavior and navigation indicators are generally consistent with the scene, the explanations occasionally misidentify maneuver intentions. This may reflect teacher-forced explanation supervision, where the model learns to describe actions from scene context rather than explicitly grounding their causes.

\paragraph{Quantitative Results}
On in-house validation scenes, GridDriver with explicit ego state, without ego state, and a constant-velocity baseline achieves 0.22/0.34/0.52m L2 error ($std\leq\pm0.01$) and 0.31/0.59/0.59\% collision rates, respectively. On 177 unseen scenes geofenced from train and val, L2 remains similar for both, ego and no-ego variants (0.22/0.36m), indicating generalization beyond the validation split. Collision rates are unavailable due to missing 3D annotations.
Perception remains challenging under domain shift. Fine-tuning BEVFormer from the official checkpoint to our setting improves car precision/recall from 0.32/0.13 to 0.56/0.62 and pedestrian precision/recall from 0.26/0.10 to 0.48/0.53. Limited in-house training data constrains adaptation.

Despite ego-state improving nuScenes metrics, qualitative observations sometimes favor the no-ego variant. Overall, results demonstrate promising cross-domain transfer while highlighting remaining real-world limitations.

\subsection{Language Influence on Planning}
\label{sec:language-intervention}

\begin{figure}[t]
    \centering
    \resizebox{\columnwidth}{!}{%
        \includegraphics[width=0.9\columnwidth]{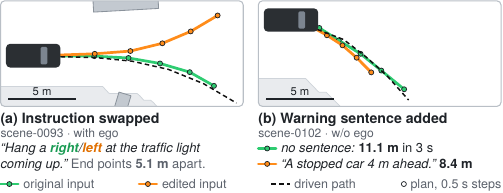}%
    }   
    \caption{Model predictions on (a) different navigation instructions and (b) injected counterfactual reasoning on nuScenes.}
    \label{fig:nuscenes-cf}
\end{figure}

\begin{table}[t]
\centering
\caption{Language intervention effects on nuScenes 3s plans (m). $N$: frames; $\Delta d$: path-length change; $\Delta y$: lateral shift (turn-aligned for instruction edits). Bold indicates statistical significance at the 95\% confidence level. With/no ego: with or without explicit ego-state input.}
\label{tab:steer}

\begingroup
\footnotesize
\setlength{\tabcolsep}{3pt}
\renewcommand{\arraystretch}{1.05}

\begin{tabular*}{\columnwidth}{
@{\extracolsep{\fill}}lrrrrr@{}}
\toprule
& & \multicolumn{2}{c}{With ego}
    & \multicolumn{2}{c}{No ego} \\
\cmidrule(lr){3-4}\cmidrule(l){5-6}
Language intervention & $N$
    & $\Delta d$ & $\Delta y$
    & $\Delta d$ & $\Delta y$ \\
\midrule
\multicolumn{6}{@{}l}{\textit{Explanation edits}} \\
Remove sentence
    & 5,119 & \textbf{-0.10} & \textbf{+0.01}
    & \textbf{-0.24} & 0.00 \\
Generic text while moving
    & 2,691 & \textbf{+0.07} & 0.00
    & \textbf{+0.14} & \textbf{-0.01} \\
\quad while stopped
    & 879 & \textbf{+0.96} & -0.01
    & \textbf{+1.10} & +0.01 \\
Hazard: Insert pedestrian warning
    & 2,691 & \textbf{-0.48} & \textbf{+0.05}
    & \textbf{-0.75} & \textbf{+0.06} \\
\quad pedestrian present
    & 578 & \textbf{-0.46} & \textbf{+0.05}
    & \textbf{-0.78} & \textbf{+0.05} \\
\quad pedestrian absent
    & 2,113 & \textbf{-0.49} & \textbf{+0.05}
    & \textbf{-0.74} & \textbf{+0.06} \\
Hazard: Insert stopped-car warning
    & 2,691 & \textbf{-0.56} & \textbf{+0.09}
    & \textbf{-0.84} & \textbf{+0.04} \\
Clearance: "Road ahead is free"
    & 879 & \textbf{+0.60} & \textbf{+0.02}
    & \textbf{+0.31} & 0.00 \\
\quad while road free
    & 493 & \textbf{+0.64} & \textbf{+0.03}
    & \textbf{+0.39} & 0.00 \\
\quad while road blocked
    & 386 & \textbf{+0.55} & 0.00
    & \textbf{+0.21} & 0.00 \\
\midrule
\multicolumn{6}{@{}l}{\textit{Navigation edits}} \\
Turn $\rightarrow$ opposite
    & 1,003 & -0.07 & \textbf{+0.65}
    & \textbf{-0.18} & \textbf{+0.52} \\
Turn $\rightarrow$ straight
    & 1,003 & \textbf{+0.43} & -0.04
    & \textbf{+0.35} & -0.06 \\
False turn (no junction)
    & 1,367 & 0.00 & \textbf{+0.13}
    & +0.11 & \textbf{+0.12} \\
\bottomrule
\end{tabular*}
\endgroup
\end{table}

We assess the influence of language on planning by modifying text inputs in Table~\ref{tab:steer} while keeping the nuScenes frames and model checkpoint fixed. We compare the resulting trajectories against those obtained with no text or length-matched neutral text. At the 3s horizon, $\Delta d$ denotes the change in predicted path length (positive = longer), while $\Delta y$ measures the lateral shift relative to the control trajectory. For navigation edits, positive $\Delta y$ indicates a shift toward the requested direction. For turn-to-straight edits, lateral shifts are signed relative to the original turn direction. For explanation edits, positive $\Delta y$ indicates a shift to the left in the ego frame.

The numbers confirm that pre-plan reasoning actively steers the trajectory. Removing or replacing explanations alters planned motion to a shorter vs. longer plan, particularly when the vehicle is waiting. We hypothesize that the model associates explanatory sentences with the onset of movement, as such sentences often follow a cue to proceed. The effects are similar across models with and without explicit ego-state input.
Hazard warnings shorten planned travel even when no hazard is present, whereas clearance messages increase it even when the path is blocked. Clearance examples highlight the risk of language dominating visual evidence, motivating stronger visual grounding.

Navigation instructions also steer the plan: Reversing a turn shifts the trajectory toward the requested direction, while replacing a turn with straight increases planned travel. Impossible turns leave travel distance largely unchanged but induce small lateral shifts, suggesting that scene geometry moderates, but does not prevent, responses to invalid instructions. Figure~\ref{fig:nuscenes-cf} shows two examples.

\subsection{Architecture Ablations}
\label{sec:ablations}
We conduct ablations on nuScenes without explicit LLM ego-state input, while retaining BEVFormer's ego-motion alignment and CAN-bus conditioning.
We replace 32 Q-Former queries with $18\times18$ spatial grid tokens and use scene-disjoint linear probes to examine whether BEV layout remains accessible throughout the LLM and reaches the waypoint head. 

Scene-disjoint ridge probes (three seeds) in Table~\ref{tab:qformer-s42} show that grid tokens retain more linearly decodable vehicle occupancy and drivable area than Q-Former queries ($R^2$: 0.37 vs.\ 0.10 and 0.62 vs.\ 0.32 at LLM input.) The advantage persists through the LLM, including at the plan token (0.08 vs. 0.03 and 0.30 vs. 0.15). With 28× fewer tokenizer parameters (9.7M vs. 270M), grid tokens preserve $3.7\times$ higher vehicle-occupancy $R^2$ and have faster inference (59ms vs 87ms at batch 1 on an A40 GPU). 
Grid tokens also improve lead-vehicle recall in generated explanations from 49\% to 64\% at similar precision (71\% vs. 73\%) and reduce distance error from 4.1m to 3.0m. Under identical training settings, grid tokens reduce planning L2 from $0.50\pm0.01$\,m to $0.47\pm0.00$\,m  with unchanged collision rates (0.15\%).

\begin{table}[t]
\centering
\caption{Scene-disjoint linear-probe $R^2$ for vehicle occupancy and drivable area (mean$\pm$std., 3 seeds). Comparison of Q-Former and spatial grid representations across LLM layers.}
\label{tab:qformer-s42}
\normalsize
\resizebox{\columnwidth}{!}{%
\begin{tabular}{lcccc}
\toprule
 & \multicolumn{2}{c}{Vehicle occupancy} & \multicolumn{2}{c}{Drivable area} \\
\cmidrule(lr){2-3}\cmidrule(lr){4-5}
Stage & Q-Former & Grid & Q-Former & Grid \\
\midrule
BEV features          & \multicolumn{2}{c}{0.35\std{0.00}} & \multicolumn{2}{c}{0.60\std{0.00}} \\
LLM input (tokens)    & 0.10\std{0.01} & \textbf{0.37}\std{0.00} & 0.32\std{0.00} & \textbf{0.62}\std{0.00} \\
LLM layer 14 (tokens) & 0.04\std{0.00} & \textbf{0.31}\std{0.02} & 0.19\std{0.01} & \textbf{0.59}\std{0.01} \\
LLM layer 28 (tokens) & 0.02\std{0.01} & \textbf{0.18}\std{0.01} & 0.14\std{0.02} & \textbf{0.52}\std{0.01} \\
Plan token (layer 28) & 0.03\std{0.01} & \textbf{0.08}\std{0.00} & 0.15\std{0.01} & \textbf{0.30}\std{0.01} \\
\bottomrule
\end{tabular}%
}
\end{table}

\section{On-Board Integration and Evaluation}

\subsection{Real-time Inference and Scheduling}
\label{sec:inference}

\begin{figure}[t]
    \centering
    \resizebox{\columnwidth}{!}{%
        \includegraphics[width=0.9\columnwidth]{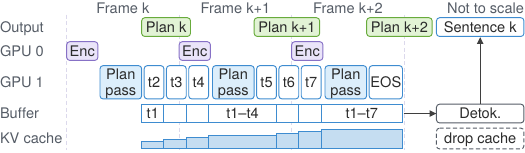}%
    }   
    \caption{Illustration of frame-wise scheduling of trajectory prediction and asynchronous language generation across two GPUs on our research vehicle (not to scale).}
    \label{fig:inference-scheduling}
\end{figure}

Figure~\ref{fig:inference-scheduling} illustrates the schedule, which interleaves token generation with subsequent planning passes to avoid delaying trajectory prediction, completing sentence $k$ in frame $k+2$ on 10Hz sensor inputs. For each frame, the planning pass produces waypoints, the first sentence token, and a KV cache. Subsequent single-token passes extend the sentence across frames while the next frame's BEV encoder runs on a separate GPU: while the planner computes the plan and continues the explanation for frame $t$, the encoder already processes frame $t{+}1$. This overlap affects only the pipeline throughput. The latencies in Table~\ref{tab:latency_ablation} are measured on a single GPU without it.
Once EOS is reached, the sentence is published with its originating frame's timestamp and the cache is discarded. 
Table~\ref{tab:latency_ablation} reports encoder and planner latency on an RTX~4090. Planning and reasoning share one forward pass, yielding the first explanation token 0.1 ms after the plan. Subsequent tokens use Key Value caching and take approximately 12\,ms each, with ten tokens per sentence on average. LoRA merging, FP16, and CUDA graphs reduce planning latency from 56.0 to 32.7\,ms (30.6\,Hz), with 153 ms mean sentence generation, providing substantial computational headroom for 10\,Hz planning.
The optimizations preserve output fidelity over 4,405 unseen in-house frames: Relative to the model as trained, ADE (0.443\,m) and FDE (0.948\,m) change by less than 0.2\,mm, maximum waypoint deviation is 3.2 cm, and maneuver/phase agreement reaches 99.8\%. All 237 compared explanations remain identical. The integrated ROS 2 pipeline achieves 9.4 plans/s, with remaining overhead from sensor transport, preprocessing, synchronization, and scheduling, which requires further system-level optimization.

\begin{table}[t]
\vspace{1.2mm}
    \centering
\caption{Inference latency (ms) on RTX~4090.
Median values except full-sentence mean.}
    \label{tab:latency_ablation}
    \resizebox{\columnwidth}{!}{%
        \begin{tabular}{lccccc}
            \toprule
            Config. & Enc. & LLM & Plan & 1st tok. & Sent. \\
            \midrule
            LoRA merged   & 38.4 & 17.6 & 56.0 & 56.1 & 168 \\
            + FP16        & 22.5 & 16.4 & 38.9 & 39.0 & 156 \\
            + CUDA graphs & \textbf{17.2} & \textbf{15.5} & \textbf{32.7} & \textbf{32.8} & \textbf{153}  \\
            \bottomrule
        \end{tabular}%
    }
\end{table}

\subsection{Live On-Board Demonstration}

\begin{figure}[t]
\centering
\begin{minipage}[t]{0.49\columnwidth}
\vspace{0pt}
\centering
\includegraphics[width=\linewidth]{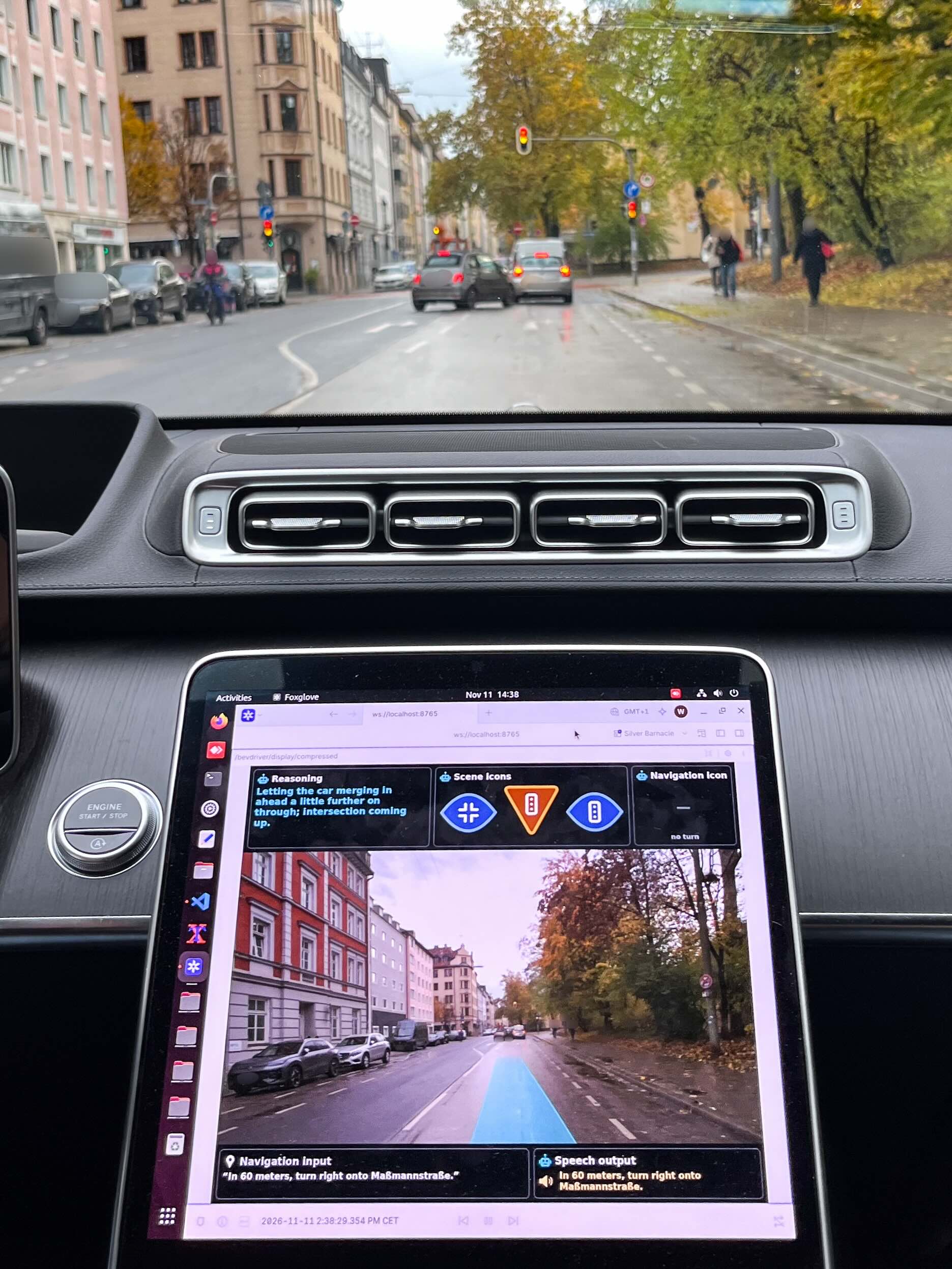}
\par\smallskip
\colorbox{gridblue}{%
\parbox{\dimexpr\linewidth-2\fboxsep\relax}{%
\raggedright\footnotesize
\parbox[t][0.85cm][t]{\dimexpr\linewidth-2\fboxsep\relax}{%
Reasoning: \emph{``Letting the car merging in ahead a little further on through; intersection coming up.''}\par}
}}
\end{minipage}\hfill%
\begin{minipage}[t]{0.49\columnwidth}
\vspace{0pt}
\centering
\includegraphics[width=\linewidth]{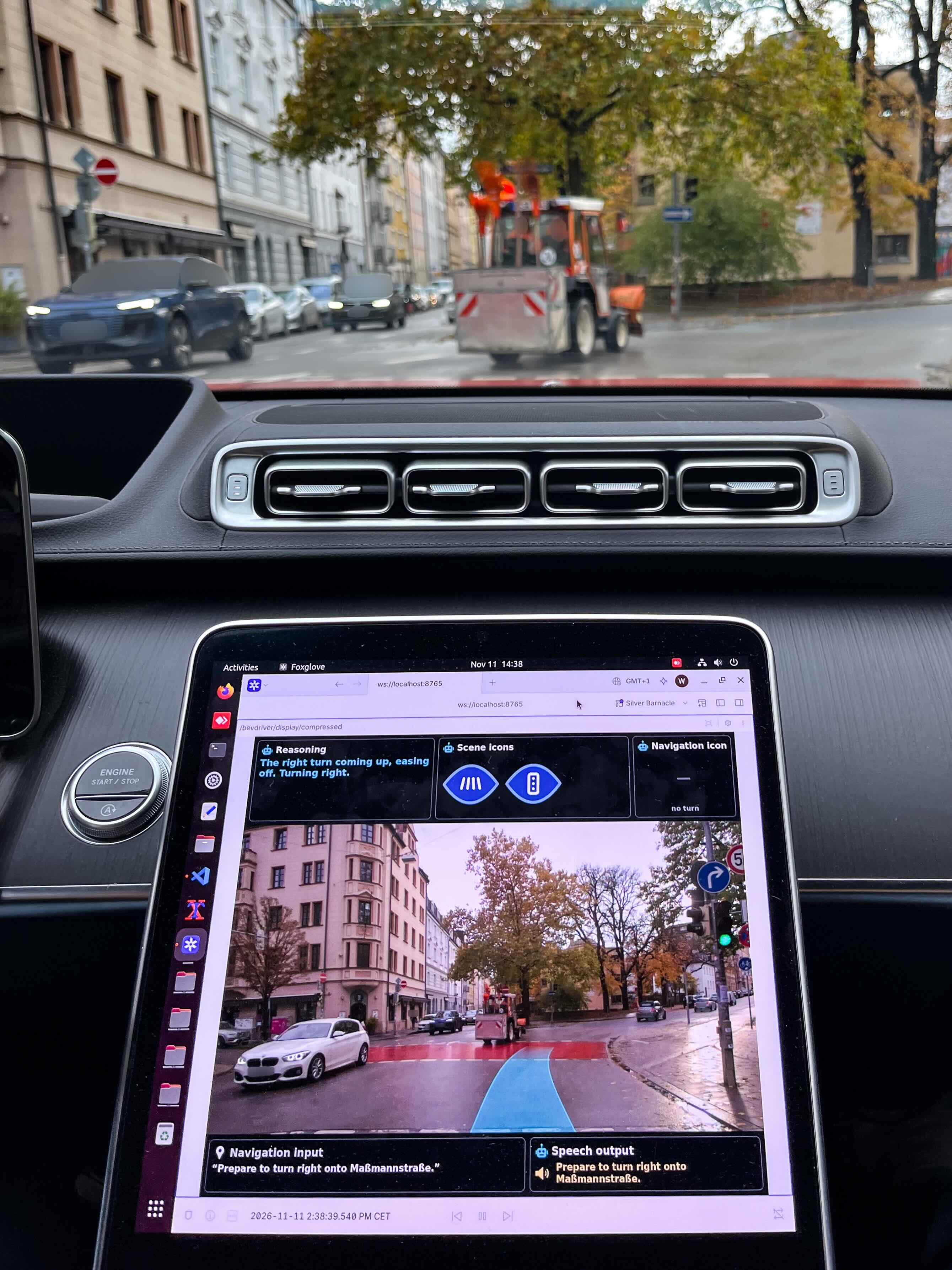}
\par\smallskip
\colorbox{gridblue}{%
\parbox{\dimexpr\linewidth-2\fboxsep\relax}{%
\raggedright\footnotesize
\parbox[t][0.85cm][t]{\dimexpr\linewidth-2\fboxsep\relax}{%
Reasoning: \emph{``The right turn coming up, easing off. Turning right.''}\par}
}}
\end{minipage}
\caption{Live urban driving showing onboard visualization and planned trajectories. Generated reasoning below each image.}
\label{fig:car_setup_live}
\end{figure}

We evaluate GridDriver in open-loop mode during real-world driving. As shown in Fig.~\ref{fig:car_setup_live}, the in-vehicle interface continuously overlays predicted trajectories onto the front-camera view achieving real-time updates. Frequent reasoning output and too many icons overwhelm the passenger, thus plan tokens are visualized as intuitive icons indicating detected road users, traffic infrastructure, and driving intentions, such as yielding or following. While some context is expected immediately (e.g. traffic light turning green), repeated explanations of a longer maneuver should be avoided. 
The model's textual reasoning is displayed upon completion and, together with Mapbox-triggered navigation instructions, communicated through synthesized speech. This setup highlights the challenge of providing correct, timely, contextually relevant, and appropriately detailed reasoning during real-world driving, where delayed or excessive explanations may reduce their usefulness to passengers.
 
\section{Discussion}
\label{sec:discussion}

GridDriver achieves competitive BEV-based language-conditioned planning, while spatial grid tokens enable more accurate linear probing of scene information than Q-Former compression. However, coarse BEV resolution (2.048 m/cell), limited coverage (±51.2 m), and imperfect cross-domain perception constrain scene understanding. Low trajectory error also does not guarantee suitable driving behavior, motivating complementary safety, progress, and comfort evaluation.

Language interventions reveal that explanations influence trajectories, even when they contradict visual scene evidence. Shared planning features couple trajectories and structured indicators, but free-form explanations may not faithfully represent driving decisions. 
Real-world shadow-mode operation does not steer the vehicle but demonstrates computational feasibility. It also enables evaluation of system requirements and passenger interactions under real driving conditions, without establishing closed-loop safety. Future work includes closed-loop validation, improved visual grounding, more diverse navigation instructions, and onboard question answering. Reinforcement learning and improved reasoning methods may further enhance planning robustness and causal understanding in diverse driving scenarios.

\section{Conclusion}
\label{sec:conclusion}

We presented GridDriver, a BEV-conditioned language planner that jointly predicts trajectories, driving indicators, and language explanations. The model achieves competitive planning and QA performance on public benchmarks, while real-time open-loop deployment on a research vehicle demonstrates onboard feasibility. Language interventions and real-world observations reveal both the influence of language on planning and remaining challenges in visual grounding and reasoning consistency. Code and reproducibility resources will support further research on onboard BEV–based LLM planning.

\section*{ACKNOWLEDGMENT}
The research leading to these results is funded by the German Federal Ministry for Economic Affairs and Energy (BMWE) within the project “NXT GEN AI METHODS --Generative Methoden für Perzeption, Prädiktion und Planung” (grant no. 19A23914M). The authors gratefully acknowledge the scientific support and HPC resources provided by the Erlangen National High Performance Computing Center (NHR@FAU) of the Friedrich-Alexander-Universität Erlangen-Nürnberg (FAU). The authors are solely responsible for the content of this publication.

AI Use Disclosure: AI assisted with experiment orchestration, statistical analysis, and grammar editing. All reported results were derived from actual experimental runs and verified by the authors, who take full responsibility for the final manuscript.

\bibliographystyle{IEEEtran}
\bibliography{refs}

@inproceedings{winter2025bevdriver,
  title = {BEVDriver: Leveraging BEV Maps in LLMs for Robust Closed-Loop Driving},
  author = {Winter, Katharina and Azer, Mark and Flohr, Fabian B.},
  booktitle = {\IROS},
  pages = {20379--20385},
  year = {2025},
  month = oct,
}

@inproceedings{shao2024lmdrive,
  title = {{LMDrive}: Closed-Loop End-to-End Driving with Large Language Models},
  author = {Shao, Hao and others},
  booktitle = {\CVPR},
  pages = {15120--15130},
  year = {2024},
  month = jun,
}

@inproceedings{fsdrive,
  title = {{FutureSightDrive}: Thinking Visually with Spatio-Temporal {CoT} for Autonomous Driving},
  author = {Shuang Zeng and others},
  booktitle = {\NIPS},
  pages = {67299–-67318},
  year = {2025},
  month = dec,
}

@article{emma,
  title = {{EMMA}: End-to-End Multimodal Model for Autonomous Driving},
  author = {Jyh-Jing Hwang and others},
  journal = {\TMLR},
  pages = {},
  year = {2025},
  month = jul,
}

@inproceedings{omnidrive,
  title = {{OmniDrive}: A Holistic Vision-Language Dataset for Autonomous Driving with Counterfactual Reasoning},
  author = {Shihao Wang and others},
  booktitle = {\CVPR},
  pages = {22442--22452},
  year = {2025},
  month = jun,
}

@inproceedings{drivevlm,
  title = {{DriveVLM}: The Convergence of Autonomous Driving and Large Vision-Language Models},
  author = {Xiaoyu Tian and others},
  booktitle = {\CORL},
  pages = {4698--4726},
  year = {2025},
  month = nov,
}

@inproceedings{gptdriver,
  title = {{GPT-Driver}: Learning to Drive with {GPT}},
  author = {Jiageng Mao and Yuxi Qian and Junjie Ye and Hang Zhao and Yue Wang},
  booktitle = {\NEURIPSW},
  pages = {},
  year = {2023},
  month = dec,
}

@inproceedings{zhou2026opendrivevla,
  title = {{OpenDriveVLA}: Towards End-to-End Autonomous Driving with Large Vision Language Action Model},
  author = {Zhou, Xingcheng and Han, Xuyuan and Yang, Feng and Ma, Yunpu and Tresp, Volker and Knoll, Alois},
  booktitle = {\AAAI},
  pages = {13782--13790},
  year = {2026},
  month = jan,
}

@inproceedings{renz2025simlingo,
  title = {{SimLingo}: Vision-Only Closed-Loop Autonomous Driving with Language-Action Alignment},
  author = {Renz, Katrin and Chen, Long and Arani, Elahe and Sinavski, Oleg},
  booktitle = {\CVPR},
  pages = {11993--12003},
  year = {2025},
  month = jun,
}

@inproceedings{zhou2026autovla,
  title = {{AutoVLA}: A Vision-Language-Action Model for End-to-End Autonomous Driving with Adaptive Reasoning and Reinforcement Fine-Tuning},
  author = {Zhou, Zewei and others},
  booktitle = {\NIPS},
  pages = {27920-–27956},
  year = {2025},
  month = dec,
}

@inproceedings{bevformer,
  title = {{BEVFormer}: Learning Bird's-Eye-View Representation from Multi-Camera Images via Spatiotemporal Transformers},
  author = {Zhiqi Li and others},
  booktitle = {\ECCV},
  pages = {1--18},
  year = {2022},
  month = oct,
}

@inproceedings{navsim,
  title = {{NAVSIM}: Data-Driven Non-Reactive Autonomous Vehicle Simulation and Benchmarking},
  author = {Daniel Dauner and others},
  booktitle = {\NIPS},
  pages = {28706--28719},
  year = {2024},
  month = dec,
}

@inproceedings{li2026sgdrive,
  title = {{SGDrive}: Scene-to-Goal Hierarchical World Cognition for Autonomous Driving},
  author = {Li, Jingyu and others},
  booktitle = {\CVPR},
  pages = {4032--4042},
  year = {2026},
  month = jun,
}

@inproceedings{park2026driveik,
  title = {Grounding Driving VLA via Inverse Kinematics},
  author = {Park, Junsung and Shim, Hyunjung},
  booktitle = {\NIPS},
  pages = {},
  year = {2026},
  month = dec,
}

@article{wayve2024lingo2,
  title = {{LINGO-2}: Driving with Language},
  author = {{Wayve}},
  journal = {Wayve Research Blog},
  pages = {},
  year = {2024},
  month = apr,
}

@article{nvidia2025alpamayo,
  title = {{Alpamayo-R1}: Bridging Reasoning and Action Prediction for Generalizable Autonomous Driving in the Long Tail},
  author = {Wang, Yan and others},
  journal = {arXiv preprint arXiv:2511.00088},
  pages = {},
  year = {2025},
  month = oct,
}

@inproceedings{cui2025onboard,
  title = {On-Board Vision-Language Models ({VLMs}) for Personalized Motion Control of Autonomous Vehicles},
  author = {Cui, Can and others},
  booktitle = {\IROS},
  pages = {20349--20356},
  year = {2025},
  month = oct,
}

@inproceedings{luo2025whatshappening,
  title = {“What's Happening”-A Human-centered Multimodal Interpreter Explaining the Actions of Autonomous Vehicles},
  author = {Luo, Xuewen and Ding, Fan and Chen, Ruiqi and Panda, Rishikesh and Loo, Junnyong and Zhang, Shuyun},
  booktitle = {\WACVW},
  pages = {1073--1080},
  year = {2025},
  month = feb,
}

@inproceedings{drivepi,
  title = {{DrivePI}: Spatial-aware 4D MLLM for Unified Autonomous Driving Understanding, Perception, Prediction and Planning},
  author = {Liu, Zhe and others},
  booktitle = {\CVPR},
  pages = {3688--3698},
  year = {2026},
  month = jun,
}

@inproceedings{rdadriver,
  title = {Making large language models better planners with reasoning-decision alignment},
  author = {Huang, Zhijian and others},
  booktitle = {\ECCV},
  pages = {73--90},
  year = {2024},
  month = oct,
}

@inproceedings{cao2025pseudo,
  title={Pseudo-Simulation for Autonomous Driving},
  author={Cao, Wei and others},
  booktitle={\CORL},
  year={2025},
  pages={4709--4722},
  month=sep,
}

@inproceedings{li2025recogdrive,
  title = {{ReCogDrive}: A Reinforced Cognitive Framework for End-to-End Autonomous Driving},
  author = {Yongkang Li and others},
  booktitle = {\ICLR},
  pages = {157518--157556},
  year = {2026},
  month = apr,
}

@inproceedings{liao2025diffusiondrive,
  title = {{DiffusionDrive}: Truncated Diffusion Model for End-to-End Autonomous Driving},
  author = {Bencheng Liao and others},
  booktitle = {\CVPR},
  pages = {12037--12047},
  year = {2025},
  month = jun,
}

@inproceedings{li2026drivevlaw0,
  title = {{DriveVLA-W0}: World Models Amplify Data Scaling Law in Autonomous Driving},
  author = {Li, Yingyan and others},
  booktitle = {\ICLR},
  pages ={7890--7911},
  year = {2026},
  month = apr,
}

@article{zhao2026simwam,
  title   = {SimWAM: A Simple World Action Model for End-to-End Autonomous Driving},
  author  = {Zhao, Zongchuang and others},
  journal = {arXiv preprint arXiv:2608.07468},
  year    = {2026}
}

@inproceedings{li2024ego,
  title={Is ego status all you need for open-loop end-to-end autonomous driving?},
  author={Li, Zhiqi and others},
  booktitle={\CVPR},
  pages={14864--14873},
  year={2024}
}

@inproceedings{chen2024asynchronous,
  title={Asynchronous large language model enhanced planner for autonomous driving},
  author={Chen, Yuan and Ding, Zi-han and Wang, Ziqin and Wang, Yan and Zhang, Lijun and Liu, Si},
  booktitle={\EECV},
  pages={22--38},
  year={2024},
  month=Oct,
}

@misc{li2026oxygen,
  author = {Xiangyu Li and Huaizhi Tang and Xin Ding
            and Weijun Wang and Ting Cao and Yunxin Liu},
  title = {{OxyGen}: Unified KV Cache Management
           for VLA Inference under Multi-Task Parallelism},
  year = {2026},
  month = mar,
  journal = {arXiv preprint arXiv:2603.14371},
  note = {Accepted to NeurIPS 2026}
}

\end{document}